# Semantic Image Networks for Human Action Recognition

Sunder Ali Khowaja and Seok-Lyong Lee

Department of Industrial and Management Engineering, Hankuk University of Foreign Studies, Global Campus, Yongin, South Korea.

Sandar.ali@usindh.edu.pk, sllee@hufs.ac.kr

Abstract

In this paper, we propose the use of a semantic image, an improved representation for video analysis, principally in combination with Inception networks. The semantic image is obtained by applying localized sparse segmentation using global clustering (LSSGC) prior to the approximate rank pooling which summarizes the motion characteristics in single or multiple images. It incorporates the background information by overlaying a static background from the window onto the subsequent segmented frames. The idea is to improve the action-motion dynamics by focusing on the region which is important for action recognition and encoding the temporal variances using the frame ranking method. We also propose the sequential combination of Inception-ResNetv2 and long-short-term memory network (LSTM) to leverage the temporal variances for improved recognition performance. Extensive analysis has been carried out on UCF101 and HMDB51 datasets which are widely used in action recognition studies. We show that (i) the semantic image generates better activations and converges faster than its original variant, (ii) using segmentation prior to approximate rank pooling yields better recognition performance, (iii) The use of LSTM leverages the temporal variance information from approximate rank pooling to model the action behavior better than the base network, (iv) the proposed representations can be adaptive as they can be used with existing methods such as temporal segment networks to improve the recognition performance, and (v) our proposed four-stream network architecture comprising of semantic images and semantic optical flows achieves state-of-the-art performance, 95.9% and 73.5% recognition accuracy on UCF101 and HMDB51, respectively.

Keywords: Motion representation, semantic information, convolutional neural networks, human action recognition, long-short term memory networks, frame ranking.

1. Introduction

Human action recognition from videos has been an active research area due to its variety of applications such as surveillance, healthcare, robotics, and so forth. The performance of action recognition has drastically been improved in recent years mainly due to deep network architectures. However, the optimal representation of the videos is still an ongoing research issue. In the last decade, researchers have proposed local spatiotemporal descriptors based on motion, gradient, dense trajectories, dense sampling, and spatiotemporal interest points (Jain et al. 2013; Laptev 2005; H. Wang et al. 2013) for action recognition. Furthermore, the advancement in object recognition techniques motivated the research community to combine these descriptors with well-developed encoding schemes like fisher vectors (H. Wang and Schmid 2013).

Many recent works rely on the contents in image sequences rather than modeling their dynamics. These contents are understood from the images which are then fed to the discriminative learning methods as their input. Thanks to the convolutional and recurrent neural networks (RNNs), the features are learned in an end to end manner to solve specific problems such as action recognition. However, using these discriminative learning methods does not solve the problem better representations as they are largely based on spatiotemporal filters for classifying actions. Earlier works focused on the representation such as dynamic textures introduced by (Doretto et al. 2003) and flow-based appearance proposed by (H. Wang and Schmid 2013). (Fernando et al. 2015) proposed a way to represent the motion dynamics of image sequences by their temporal order. These representations were successful in improving the performance of an action recognition system. (Bilen et al. 2016, 2018) extended the idea to construct a dynamic image using the temporal order of image sequences. The idea was to incorporate the motion dynamics into a single image. The image can be fed to any standard convolutional neural network (CNN) for an end to end learning. All the existing representations have a common goal: modeling the action-motion dynamics to improve the performance. The CNNs extract the features such as spatiotemporal, edge, and gradient features based on the progression of convolutional layers. In this work, we extend the notion of Bilen et al. by using the temporal order of image sequences after performing the segmentation and incorporating semantics, i.e. background information; we refer to this representation as a semantic image (SemI). The idea is to improve the action-motion dynamics by providing frames with sparse characteristics such that the region of interest should be focused more rather than the whole image. It is proved that incorporating background information can have a positive influence in recognition performance (Joe Yue-Hei Ng et al. 2015). In this regard, we overlay a static background on all the subsequent segmented frames to impose the semantic information in a single SemI. It is apparent that in video analysis, the background changes frequently. In this regard, we divide the whole video into windows of overlapping frames, and a static background from each subset of frames is overlaid to the subsequent segmented frames in that window. In this way, we incorporate the changing background information in multiple SemI. The SemI improves the video representation in terms of compactness, flexibility, effectiveness, and adaptability. The compactness refers to the summarization of motions from all video frames on to a single image along with semantic information. The flexibility represents the use of the representation throughout the network architecture. The effectiveness is the ability to handle back-propagation for end-to-end learning. The SemI can also be used with existing network architectures which shows the adaptability trait.

It is apparent that the representations play an important part in the recognition pipeline, but without a good learning method, the representations cannot achieve their true potential. Existing works mainly use CNN, RNN or LSTM which is a variant of RNN. Both of the networks, CNN and RNN., have their own advantages and disadvantages. For instance, RNNs are good at temporal modeling, but they cannot extract high-level features, while CNNs are good at extracting high-level features but do not perform well when dealing with sequential modeling. Additionally, existing works employ mostly the pre-trained networks such as CaffeNet (Jia et al. 2014), AlexNet (Krizhevsky et al. 2012), ResNet (He et al. 2016) and the similar ones, which increase the depth of the network but not the width. These pre-trained models have specific filter sizes defined in each convolutional layer. The provision of liberty to the network for selecting the filter size has a profound effect on the object recognition studies, i.e. Inception networks. However, such networks have not been adopted for action recognition studies so far.

In this work, we propose semantic image networks which uses SemI to train the SR-LSTM. We construct semantic images by performing segmentation with localized sparse segmentation using global clustering

(LSSGC) and overlaying the background information on to the subsequent segmented frames followed by the approximate rank pooling (Bilen et al. 2016). The term sparse segmentation refers to the segmentation of regions which are of interest with respect to the field-of-view. We apply the approximate rank pooling on the semantic images to summarize motions with respect to the temporal order of frames. Some examples of semantic images are shown in figure 1. It can be visualized that the semantic images not only summarize the repeated motions as shown for 'Apply Eye Makeup' but also the direction of the motion. For instance, the horizontal motion is indicated in 'Archery' and 'Boxing Punching Bag' whereas the vertical motion is indicated in 'Handstand Walking' Action.

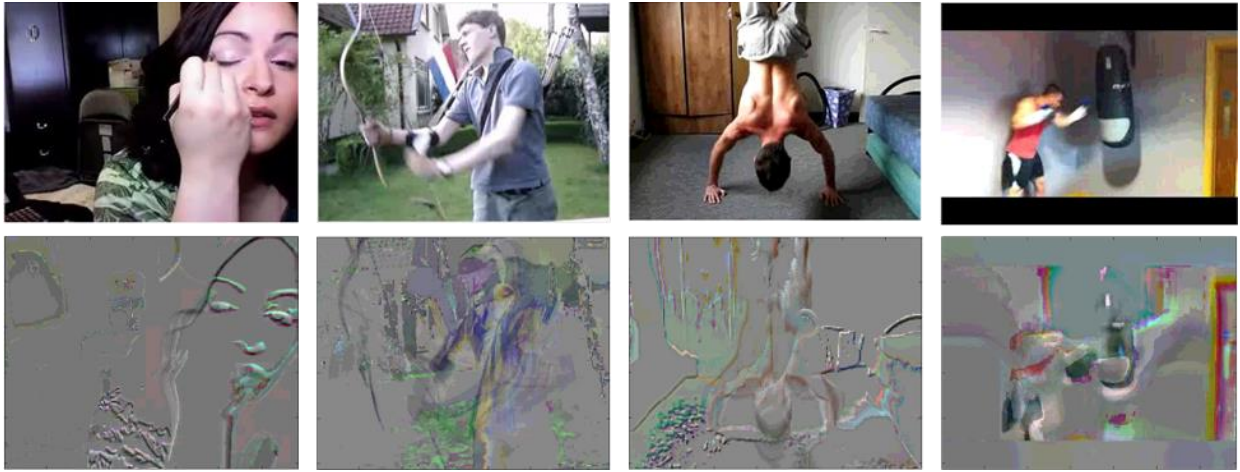

Figure 1 Semantic images derived from RGB video frames. From left to right (i) Apply Eye Makeup, (ii) Archery, (iii) Handstand Pushups, and (iv) Boxing Punching Bag. The top row shows RGB images whereas the bottom one shows semantic images

The sequential residual LSTM (SR-LSTM) network uses the pre-trained Inception-ResNetv2 (Szegedy et al. 2017) network sequenced with LSTM to perform end-to-end learning. The reason for using such architecture and configuration is three-fold. The first is to give the liberty to the network for choosing the filter size by using inception modules in the pre-trained architecture. The second is to get better performance in action recognition, and the third is to model temporal variances better for temporally ordered frames. Previous studies have tried to combine the characteristics of both CNN and LSTM which have been proved to be beneficial (Donahue et al. 2017), nevertheless modeling the temporal information from plain RGB sequences have limited benefits. In this study, we leverage the temporal variances incorporated in SemI with LSTM to handle the sequential modeling of data in a better way. The approximate rank pooling is applied to the windows of partially overlapped frames; therefore, a frame can have a high rank in the first window but may achieve low rank in the next one in case if it occurs in an overlap. This kind of phenomena is referred to as temporal variance in our study. The contributions of this study are summarized as follows:

- We propose a method (LSSGC) for dynamically segmenting the image.
- We introduce a semantic image to represent motion dynamics over a sequence of frames.
- We propose an SR-LSTM based semantic image network to leverage temporal information and variance.
- We use the semantic images with four-stream and temporal segment networks (a state-of-the-art network design) to show that the proposed representation is adaptable.
- We report state-of-the-art results using semantic image networks.

The rest of the paper is structured as follows: Section 2 consolidates the existing works carried out for action recognition. Section 3 defines the methodology for LSSGC, adding semantics to the segmentation, and approximate rank pooling for constructing semantic images. Section 4 provides details regarding SR-LSTM. The experimental results are shown in Section 5 followed by the conclusion and future work in Section 6.

2. Related Works

In this section, we provide a comprehensive review for state-of-the-art methods on action recognition using videos along with some distinction with respect to the image modalities, network characteristics, multiple streams, and long-short term dynamics.

2.1 RGB Images

The majority of the existing works for action recognition is based on the stack of still images. Researchers have used both deep learning and shallow learning methods for recognition of action using RGB images. (Fernando et al. 2015) proposed rank pooling on RGB images to explore the temporal changes in videos. The study combines the temporally aligned videos with different parametric models such as rank support vector machines (Rank SVM) and support vector regression (SVR) (Smola and Schölkopf 2004) to measure the performance of action recognition. They suggested that the effect of rank pooling can improve the performance up to 7 – 10 % in comparison to the average pooling layers. (H. Wang and Schmid 2013) proposed improved dense trajectories (IDT) from series of still images. These trajectories are computed using speed up robust features (SURF) (Willems et al. 2008) and dense optical flows combined with random sample consensus (RANSAC) (Fischler and Bolles 1981). They claimed that IDTs could help to improve many motion-based detectors. The stack of RGB images has also been used with deep learning architectures for action recognition. (Joe Yue-Hei Ng et al. 2015) used the sequence of RGB images along with the background information, suggesting that some activities are only performed at a specific place such as "hockey penalty" and "basketball" will always be played in ground and basketball court, respectively. They used convolutional neural networks (CNNs) along with the long-short-term memory networks (LSTM) to classify the action. The study achieved 88.6% accuracy on UCF101 dataset. (Simonyan and Zisserman 2014) used the combination of stacked RGB images and optical flows with CNNs to improve the action recognition performance. They reported the mean class accuracy on UCF101 and HMDB51 to be 86.9% and 58.0%, respectively. In our study, we also use RGB images but to transform them in semantic images. We also use the network based on RGB images for our two- and four-stream networks.

2.2 Motion Information

Motion information is considered to be of vital importance when classifying human actions. Methods capturing motion information from a sequence of images have been extensively used in existing studies. The techniques for summarizing motions include motion history images (MHI) and motion energy images (MEI) (Bobick and Davis 2001), dynamic textures (Kellokumpu et al. 2008), and optical flows (Ali and Shah 2010). (Bobick and Davis 2001) introduced the concept of MHI for action and motion recognition from videos. The MEIs find the regions where motion is present and highlight those image regions to show different motion patterns. MEI uses spatial motion-distribution pattern by computing the image differences and summing their squares. The MHI is the encoded version of MEI which computes the motion of each pixel at a given location. Optical flow based methods use principle components for summarizing the motion between successive frames. (Ali and Shah 2010) proposed the use of optical flows for generating kinematic features. These features are then trained using multiple instance learning for performing action recognition. (Yan Ke et al. 2005) performed action detection and recognition by computing optical flows from sub-spaces volumes of the image sequences. As discussed in the previous subsection, many two-stream networks use the optical flow images as one of the modality to train it with the combination of RGB images or other representations (Feichtenhofer, Pinz, and Zisserman 2016; Simonyan and Zisserman 2014). Another form of image representation is the dynamic image which was introduced by (Bilen et al. 2016).

The dynamic image summarizes the motion dynamics of a video in a single image. Unlike the MEI and MHI, the summarization in dynamic images is based on the ranking order of the frames proposed in (Fernando et al. 2015). The concept of ranking frames can also be applied to optical flows for generating dynamic optical flows. In our work, we also apply the approximate rank pooling for ordering the frames. However, we apply the approximate rank pooling on the segmented images which are then fused with a static background. We show that such kind of pre-processing improves the action-motion dynamics as well as the performance in terms of action recognition accuracy.

### 2.3 Spatio-temporal dynamics

The techniques using spatiotemporal dynamics are evolved from patterns extracted using sequenced data which were initially used for texture recognition (Doretto et al. 2003). These techniques take into account the sequential data and estimate the parameters of the model using an autoregressive moving average for constructing dynamic textures. (Kellokumpu et al. 2008) applied the dynamic textures for time-varying sequences to recognize human actions. The creation of dynamic textures was based on local binary patterns (LBP) (Ojala et al. 2002) descriptors which help to encode the micro-texture for the 2D neighborhood of the pixels; they referred to it as binary strings. (Le et al. 2011) extended the idea to use independent subspace analysis (ISA) with the dynamic textures to recognize human actions. The representations using ISA were extracted hierarchically to handle invariant representations which improved the accuracy of the recognition task. Nonetheless, these techniques capture the motion dynamics from an action video, but they do not consider the video sequence for modeling the motion characteristics.

### 2.4 Spatio-Temporal Volumes

In recent years, the researchers have considered the use of 3D volumes which adds a third dimension to the 2D images, i.e. time. These 3D volumes are derived from spatiotemporal templates (Pirsiavash et al. 2009; Rodriguez et al. 2008; Shechtman and Irani 2005). (Ji et al. 2013) proposed the 3D CNNs which capture the spatial as well as temporal information from the videos. The spatial information is captured using 2D, and the motion information from multiple frames is accounted for the third dimension to learn the features. (Tran et al. 2015) proposed the use of convolutional 3D features (C3D) which were learned using 3D CNNs on large-scale datasets. Their study showed that using many feature representations such as IDTs, optical flows and so forth, 3D CNNs can boost the performance. (Hara et al. 2018) recently proposed the use of 3D kernels with very deep CNNs to boost the performance of different recognition tasks. They employed ResNeXt101 using 3D kernels while experimenting with a larger number of filters, i.e. 64 and achieved better results for UCF101 and HMDB51 as compared to the filter size of 32. The problem with 3D CNNs is that they require a large amount of annotations to bring the natural representation of frame sequences. Moreover, the use of 3D filters increases the number of parameters for the employed network architecture which increases the computational complexity. An alternative way of using spatiotemporal information is to employ multi-stream networks with different modalities as performed in (Simonyan and Zisserman 2014). In this work, we too exploit the use of multiple modalities such as semantic images and semantic optical flows to extract spatiotemporal information.

### 2.5 Multi-Stream Networks

Multi-stream networks refer to the combination of single stream networks trained on a specific modality such as RGB, optical flow, and so forth while using their late fusion for drawing final classification label. The multi-stream networks have been used in a variety of domains. Earlier examples of multi-stream networks include Siamese architecture which learns to classify the input and measure the similarity between the output of the classification (Chopra et al. 2005). (Simonyan and Zisserman 2014) used the pre-computed optical flows along with the RGB images to train the two-stream CNNs to boost the recognition result. (Feichtenhofer, Pinz, and Zisserman 2016) extended the idea by fusing the architectures at various levels of CNN architecture. The fusion was based on 3D CNNs to combine different modalities for an end to end training. (Bilen et al. 2016) presented the idea of using two-stream networks on different combinations such as RGB, optical flows, dynamic images, and dynamic optical flows. In (Bilen et al. 2018) the authors extended their work by employing four-stream networks using multiple dynamic and RGB images, optical flows and dynamic optical flows with a very deep network architecture i.e. ResNeXt50 and ResNeXt101. They reported the state-of-the-art results on UCF101 and HMDB51 datasets with 95.5% and 72.5%, respectively. Similar to the existing studies, we also explore the use of two- and four-stream networks to boost our recognition performance.

2.6 Long and Short Term Dynamics
In the previous subsections, we mostly mentioned the studies which capture short-term dynamics, i.e. from smaller windows. Another category of deep architectures uses recurrent neural networks (RNN) which are capable of capturing long-term motion dynamics. (Donahue et al. 2017) proposed the use of sequential networks comprising of CNN for feature extraction and LSTM for modeling the temporal dependencies. (Srivastava et al. 2015) proposed a LSTM based autoencoder which reconstructs the next frame by taking into account the current one. Their study used the representations from the LSTM autoencoder for further classification of actions. (Ma et al. 2018) extended the work of (Simonyan and Zisserman 2014) by combining the two stream, i.e. RGB and stacked optical flows with the LSTMs to model the temporal dependencies. In our work, we also use approximate rank pooling for long-term dynamics of the video sequences and sequence LSTMs with Inception-ResNetv2 to model the temporal variance.

2.7 Other works
There are other works on action recognition which do not solely focus on the video representations or network architectures. However, they do focus on contextual information which can help to improve the action recognition task. (Jain et al. 2015) proposed the method to model human-object interaction by considering the object detection used for a particular action. They used the motion characteristics alongside the objects to recognize actions. (L. Wang et al. 2015) combined the characteristics of handcrafted features and deep learning network features to train in an end to end manner. They referred to the method as trajectory-pooled deep convolutional descriptors (TDD). (Cheron et al. 2015) proposed pose-based CNN by extracting the poses from the actors and then extracting appearance and optical flow based features from each of the body parts. The resultant normalized feature vector was then trained using SVM to predict the action label. (L. Wang, Xiong, et al. 2016) presented the good practices for training deep learning architectures on such video representations using sample-based approaches. They show that such learning strategy can boost the action recognition results. In our work, we use the semantic images and semantic optical flows using temporal segment networks (L. Wang, Xiong, et al. 2016) to prove the adaptability of the proposed representations.

3. Proposed Method
In this section, we present our methodology to construct semantic images. The proposed method heavily relies on the segmentation process. Therefore, we first explain the segmentation method followed by the process of overlaying the static background on the subsequent segmented frames. We also explain the method of approximate rank pooling for construction of Pre-SemI, SemI, and SemM, accordingly. When approximate rank pooling is only applied to the segmented images, we refer to this process as Pre-SemI as it does not add the semantic information (background) yet, while we refer to SemI when pooling is applied to the background embedded images. Semantic maps (SemM) are generated when approximate rank pooling is applied to intermediate layers of our network architecture which is to show the flexibility of the proposed representation.

### 3.1 Image Segmentation

We aim to obtain the segmented image with sparse characteristics such that the important regions are segmented for adding semantics and summarizing motion dynamics. We incorporate the global k-means algorithm (Likas et al. 2003) into the piecewise-constant active contour model (Vese and Chan 2002) and its late fusion with a color map from the Potts model in (Storath and Weinmann 2014) such that the loss of region is minimal. The global k-means algorithm starts by placing the cluster centers at arbitrary positions and move them to minimize the clustering error. This method of clustering has been proven to be effective for many applications in computer vision field — the global k-means algorithm partitions the color texture regions which are homogeneous. The quantized colors characterize the information of the color texture. The vector of quantized colors is determined using General Lloyd Algorithm (Allen and Gray 2012) which computes the distortion for centroid values of each cluster as shown in equation 1.

$$\Theta = \sum_n \sum_i \omega(i) \|x(i) - c_n\|^2 \rightarrow x(i) \in C \tag{1}$$

Where $\Theta$ refers to the distortion value computed for each pixel $i$ in the $n^{th}$ cluster $C = (c_1, \dots, c_N)$ having a specific centroid value $c_n$, where $n = 1, \dots, N$. The value of centroid will be updated based on the weight $\omega$ of the color pixel as shown in equation 2.

$$c_n = \frac{\sum \omega(i)x(i)}{\omega(i)} \rightarrow x(i) \in C \tag{2}$$

Once we obtain the centroid for each cluster, we can use the cluster values for segmentation method using Mumford-Shah energy function for the piecewise-constant case (Vese and Chan 2002) as defined in equation 3. We forward the computed centroid values to the equation such that the centroid values will replace the constant vector of averages and the class will replace the cluster label.

$$F(c, \Phi) = \sum_{1 \leq n \leq N} \int_\Omega (I(x,y) - c_n)^2 \chi_n \, dxdy + \sum_{1 \leq n \leq N} v \int_\Omega |\nabla \chi_n| \tag{3}$$

The variable $\Phi$ is the vector of level set functions, $I$ is the image function with pixel values at location $(x, y)$, and $\chi_n$ is the characteristic function for cluster $n$. The variable $v$ is a fixed parameter weight for controlling the associated energy. The second term in equation 3 refers to the edges in the total length term. However, it was found in (Vese and Chan 2002) that the characteristic function for these edges ought to change in the total length term, therefore the second term should be replaced by the Heaviside function vector i.e. $H(\Phi) = (H(\phi_1), \dots, H(\phi_m))$ which outputs 0s or 1s with respect to the level set functions, where $m$ is the number of level set functions. In this regard, equation 3 can be re-written as:

$$F(c, \Phi) = \sum_{1 \leq n \leq N} \int_\Omega (I(x,y) - c_n)^2 \chi_n \, dxdy + \sum_{1 \leq m \leq M} v \int_\Omega |\nabla H(\phi_m)| \tag{4}$$

In (Vese and Chan 2002) the clusters were referred to the average constant values and can be considered as initial contour values. We know for $n$ clusters there are $2^m$ level set functions which will perform the level set evolution for the given contour. In the proposed method, we want to make the contour points adaptive and make the region larger as the epochs continue. In simple words, we will determine the initial contour points by using global k-means clustering through which the evolution of the piecewise constant model will be performed for $j$ iterations. The global k-means clustering method will repeat for $z$ epochs by decreasing the clusters and the level set functions along with the increase of region size. The output of segmentation will then be added to the Potts Model label map for obtaining the final segmented image. The modified piecewise constant model for one channel is shown in equation (5)

$$F_k(c, \Phi) = \int_{\Omega_1} (I_1(x,y) - c_1)^2 H(\phi_1) H(\phi_2) \dots H(\phi_m) \, dxdy$$

$$+ \int_{\Omega_2} (I_2(x,y) - c_2)^2 H(\phi_1) H(\phi_2) \dots H(\phi_m) \, dxdy$$

...

$$+ \int_{\Omega_{|k|}} (I_n(x,y) - c_n)^2 H(\phi_1) H(\phi_2) \ldots H(\phi_m) \, dx dy \tag{5}$$

Where $c = (c_1, \ldots, c_n)$ are the cluster means assigned using global k-means clustering and $\Phi = (\phi_1, \ldots, \phi_m)$ such that $m = \log(n)/\log(2)$. An example of global k-mean clustering for an image with 8 clusters is shown in figure 1. Considering the clusters, the equation will minimize the image with respect to the 8 cluster centers using 3 level set functions. The equation 5 will be minimized using Euler-Lagrange equations with respect to $c$ and $\Phi$ using dynamic programming scheme. Equation 5 will be applied to each of the channel, separately. The segmented images generated from each of the channel will then be combined in order to obtain the color image. The pseudocode for segmenting the image is presented in algorithm 1.

We initially selected the value of *n* to be 16 which will set the number of level sets *m = 4*. The values *z* and *a* are set to be 4 and 5, respectively. The algorithm will reduce the number of clusters by the factor of 2 which reduces the number of level sets and increases the size of the region with the progression of each epoch.

---

Algorithm 1
Input: Color Image
Output: Segmented Image
*For epochs 1: z*
   *Apply global k-means clustering*
   *Circular window size = p x p*
   *For each cluster*
     *For each window*
       *Compute the Distortion value from the circular window size*
       *Update the centroid value using the square root of Euclidean distance*
     *End For*
   *End For*
   *For channels 1:3 (color image)*
     *Assign the cluster values to c*
    *m = log(n)/log(2)*
     *For iterations 1: a*
       *Segment the image by minimizing equation 5*
       *Update the level set functions with respect to mean curvature and jump of the data*
     *End For*
   *End For*
   *Concatenate all three channels*
   *Number of Clusters = Number of Clusters/2*
   *p = p*2 – 1*
*End For*
*Temp = Apply Potts Model for vector-valued segmentation*
*Segmented Image = Element wise Addition (Temp, Segmented image map)*
*Return Segmented Image*

---

As the region size increases, we increase the circular window size accordingly which is denoted by *p* which was initially set to be 9 x 9. We refer this technique as Localized sparse segmentation using global clustering (LSSGC). The advantages of performing segmentation prior to approximate rank pooling are twofold. The first is that the focus of the CNNs would be on the action-motion dynamics due to the sparse characteristics of the image, unlike the existing ones which focus on all the pixels to represent motion (see figure 1). Since the segmented images do not exhibit different characteristics for gender and skin color, it avoids visual bias which is the second advantage of prior segmentation as shown in figure 3. For instance, in "Apply Eye Makeup" all the participants except one is black colored female, but the segmented image removes the skin color to avoid the visual bias.

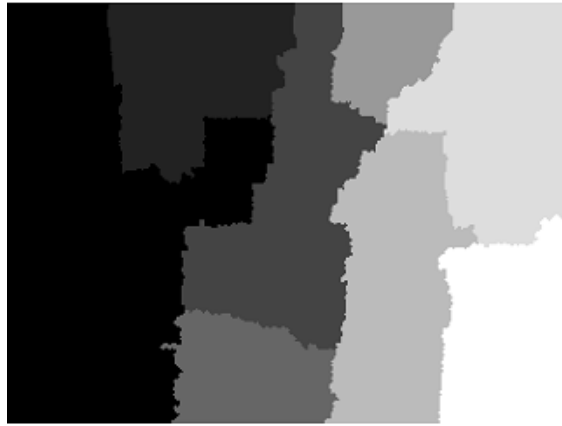

Figure 2 Example of image clusters using global k-means clustering

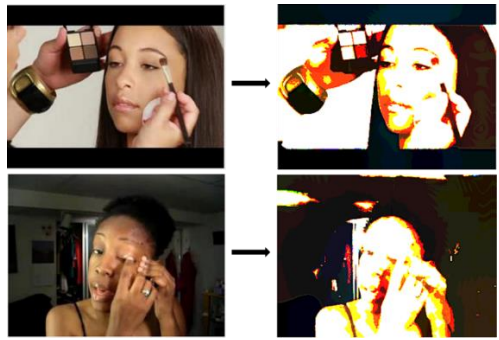

Figure 3 Example of segmented images from action 'Apply Eye Makeup' of UCF101 dataset. First row images are from subject 04 whereas the second-row images are from subject 25. It can be noticed that segmentation removes the visual bias of skin color from both subjects.

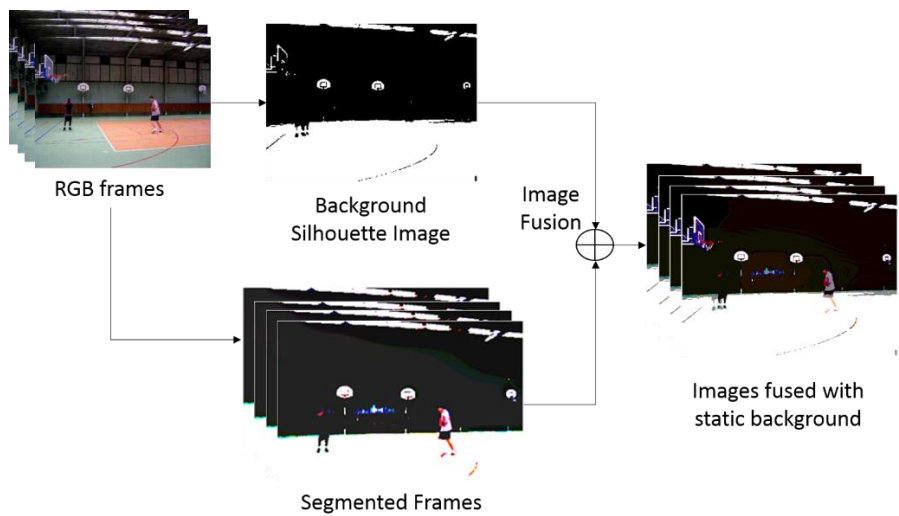

Figure 4 Process flow for adding semantic information to the segmented frames

## 3.2 Adding Semantics to the segmented image

In the related works, we referred to the study describing the importance of the background information which can help in improving the recognition results (Joe Yue-Hei Ng et al. 2015). We overlay a static background to all the subsequent frames to provide the semantic information. The process flow for overlaying background on to the segmented images is shown in figure 4, accordingly. First, the background image is estimated using the median filter method. Once the background image is computed, we generate a silhouette image by converting the logical image to black and white with image opening operation. In parallel, the frames are segmented using algorithm 1 presented in the previous sub-section. All the subsequent frames except frame 1 are then fused with the silhouette image. The fusion of images is just the overlaying of the silhouette image to all the subsequent frames based on alpha blending (Yatziv and Sapiro 2006). The fused segmented images are then fed to the approximate rank pooling for constructing SemI.

## 3.3 Approximate Rank Pooling for Semantic Images

The computation of semantic images is based on the approximate rank pooling (ARP) (Bilen et al. 2016). The ARP is based on the optimization problem which is solved by the equation 6.

$$s^* = \varsigma(I'_1, \ldots, I'_T; \psi) = argmin_s E(s)$$

where

$$E(s) = \frac{2}{T(T-1)} \sum_{q>t} max\{0, 1 - Z(q|s) + Z(t|s)\} + \frac{\beta}{2}\|s\|^2 \qquad (6)$$

The parameter vector is denoted by $s$ which defines the scores for the frames in a video. The mapping function $\varsigma(\cdot)$ maps the sequences of segmented $T$ segmented frames $I'$ to the vector $s$. The feature vector from the segmented frames is denoted by $\psi$, $Z(t|s)$ is the inner product of the parameter vector with the time average of the feature vector $\mathcal{A}_t$, i.e., $Z(t|s) = \langle s, \mathcal{A}_t \rangle$. The average feature vector is defined as $\mathcal{A}_t = \frac{1}{t}\sum \psi(I'_t)$. The constraint with respect to the variable $q$ is defined as $t < q \Rightarrow Z(t|s) < Z(q|s)$ suggesting that the larger scores are associated with the later times. The first term of the function $E(s)$ is the hinge-loss considering the incorrectly ranked pair $t < q$ based on the scores. The condition for correctly ranked pairs is $Z(t|s) + 1 < Z(q|s)$. The second term of the function $E(s)$ is the standard quadratic regularization term. In ARP the starting point is considered at $s = \vec{0}$ to obtain the first approximation using gradient descent as shown in equation 7.

$$\nabla E(\vec{0}) \propto \sum_{t<q} \nabla max\{0, 1 - Z(q|s) + Z(t|s)\}|_{s=\vec{0}}$$
$$= \sum_{t<q} \nabla \langle s, \mathcal{A}_q - \mathcal{A}_t \rangle = \sum_{t<q} \mathcal{A}_q - \mathcal{A}_t \qquad (7)$$

Further the parameter vector $s^*$ can be expanded as shown in equation 8.

$$s^* \propto \sum_{t<q} \mathcal{A}_q - \mathcal{A}_t = \sum_{t<q}\left[-\frac{1}{t}\sum_{i=1}^{t}\psi_i + \frac{1}{q}\sum_{j=1}^{q}\psi_j\right] = \sum_{t=1}^{T} \alpha_t \psi_t \qquad (8)$$

where the coefficients

$$\alpha_t = 2(1 - t + T) - (1 + T)(H_T - H_{t-1}) \qquad (9)$$

The variable $H_t$ is the $t^{th}$ harmonic number. Instead of using the coefficients of $\alpha_t$ which takes into account the intermediate average features we directly use the features $\psi_t$ for obtaining parameter vector $s^*$. From equation (7) we provide a working example as shown in equation 10.

$$\sum_{q>t} \mathcal{A}_q - \mathcal{A}_t = (\mathcal{A}_2 - \mathcal{A}_1) \ldots$$
$$(\mathcal{A}_3 - \mathcal{A}_1) + (\mathcal{A}_3 - \mathcal{A}_2) \ldots$$
$$(\mathcal{A}_T - \mathcal{A}_1) + (\mathcal{A}_T - \mathcal{A}_2) + \ldots + (\mathcal{A}_T - \mathcal{A}_{T-1}) \qquad (10)$$

One can see that each $\mathcal{A}_t$ with positive sign occurs $(t-1)$ times. For example: $\mathcal{A}_1$ occurs 0 times as positive, $\mathcal{A}_2$ occurs $(2-1)$ times as positive and so on and each $\mathcal{A}_t$ with negative sign occurs $(T-t)$ times. For example: if $T = 2$ and $t = 1$, $\mathcal{A}_1$ occurs $(2-1)$ times as negative, similarly, if $T = 4$ and $t = 1$, $\mathcal{A}_1$ occurs

(4 − 1) times as negative. Based on the harmonics the weighting function $\alpha_t$ can be written as shown in equation (11)

$$\alpha_t = (t - 1) - (\mathcal{T} - t) = t - 1 - \mathcal{T} + t = 2t - (1 + \mathcal{T}) \tag{11}$$

The reasons for using the $\psi_t$ directly for rank pooling is that the weighting function is linear with respect to $t$. For further derivations and examples refer to (Bilen et al. 2018).

### 3.4 Approximate Rank Pooling for Semantic Maps

The semantic images are constructed from the segmented frames at the input level whereas for the semantic maps the segmented images are fed to CNNs as input and the ARP is applied at intermediate layers of the same architecture. Applying the rank pooling at an intermediate level does not result in straightforward back propagation due to the dependency on the features and their intermediate averages. For intermediate layers, we can rewrite equation 6 as shown in equation 12

$$m^{(\ell)} = \varsigma\left(m_1^{\ell-1}, \dots, m_{\mathcal{T}}^{\ell-1}\right) \tag{12}$$

Where $m^{(\ell)}$ represent the feature maps on the layer $\ell$. We dropped the term $\psi$ from equation 6 as the architecture will extract the feature maps on its own. The feature maps computed at layer $(\ell - 1)$ are denoted by $\left(m_1^{(\ell-1)}, \dots, m_{\mathcal{T}}^{(\ell-1)}\right)$ for $\mathcal{T}$ image sequences. If we rewrite equation 12 with respect to the temporal average of the input patterns $\mathcal{A}_t$ it can be represented as equation 13

$$m^{(\ell)} = \sum_{t=1}^{\mathcal{T}} \alpha_t(\mathcal{A}_1, \dots, \mathcal{A}_{\mathcal{T}}) \mathcal{A}_t \tag{13}$$

Since $m^{(\ell)}$ is a linear function of the previous layer feature maps $\mathcal{A}_t$, thus, it can alternatively be written as $\left(m_1^{(\ell-1)}, \dots, m_{\mathcal{T}}^{(\ell-1)}\right) m_t^{(\ell-1)}$, substituting the values of $\mathcal{A}'s$ with $m's$, we can rewrite equation 13 as shown below

$$m^{(\ell)} = \sum_{t=1}^{\mathcal{T}} \alpha_t\left(m_1^{(\ell-1)}, \dots, m_{\mathcal{T}}^{(\ell-1)}\right) m_t^{(\ell-1)} \tag{14}$$

As suggested in (Bilen et al. 2016) due to the gradient computation of $m^{(\ell)}$ with respect to the data points $m^{(\ell-1)}$ is a challenging derivation. However, if the coefficients are independent of the features $\psi(I'_t)$ and their intermediate averages $\mathcal{A}_t$, the derivative of the ARP can be computed with the vectorized coefficients of eq. (10) as $\frac{\partial \text{vec} m^{(\ell)}}{\partial (\text{vec} m^{(\ell-1)})^{\top}} = \alpha_t \mathbb{I}$, where $\mathbb{I}$ refers to the identity matrix. This expression can be obtained by taking the derivative of eq. (13) while keeping $\alpha_t$ constant and removing its dependency on the video frames i.e. $m_t^{(\ell-1)}$ term.

### 4. Proposed Network
### 4.1 Inception-ResNetv2

Deep CNNs have been successful in achieving state-of-the-art results in many recognition domains, for instance, action recognition, object recognition and so on. Recently, Inception network architecture (Szegedy et al. 2016) has become very popular due to its improvement in the object recognition task. It has the following three distinct characteristics: the concatenating feature maps from several layers, the increase of network width along with the depth, and the reduction of the dimensions which reduces the computational cost. The problem with very deep Inception networks is the degradation phenomenon and vanishing gradients (He and Sun 2015). To overcome this problem, the ResNets were introduced (He et al. 2016), and recently the Inception-ResNetv2 (Szegedy et al. 2017) network architecture was proposed. The difference between CNN and ResNet is in the way of computing feature maps. CNN learns the feature mapping $\psi(I')$ from input $I'$, whereas the ResNets compute the function comprised of both features maps and the input i.e., $\psi(I') + I'$. This kind of learning has proven to be more accurate and faster as compared to CNNs. The Ineption-ResNetv2 achieved better results than Inception-v3 networks for object recognition, which is quite plausible and a worthy successor for exploring its performance in action recognition tasks. Inception-ResNetv2 uses cheaper inception blocks for its combination with residual networks. Furthermore, in order to match the depth of the input, a filter expansion layer which acts as a dummy layer without any subsequent activation i.e., 1 x 1 convolution is added after the inception block. This kind of dimensionality scaling is needed as inception networks reduce the dimension of the feature maps. Although the raw computation cost of Inception-ResNetv2 was the same as inceptionv4 networks, the step time of Inception-ResNetv2 is faster due to the residual addition and the reduction of layers. Another key difference which is noticed in Inception-ResNetv2 is that the batch normalization is not used on the top of the residual summations. it is stacked only on the traditional layers for reducing the overall computation cost. We used the Inception-ResNetv2 pre-trained network for single SemI and single SemM as shown in figure 5. In order to extract the SemM, we consider three intermediate levels for constructing single SemM. The ResNet Inception Branching number (RIPB) is the branching scenario for different intermediate levels. We do not construct the SemMs on each branching factor at once. Rather we repeat the experiments for one of the branching factors at a time to evaluate the performance.

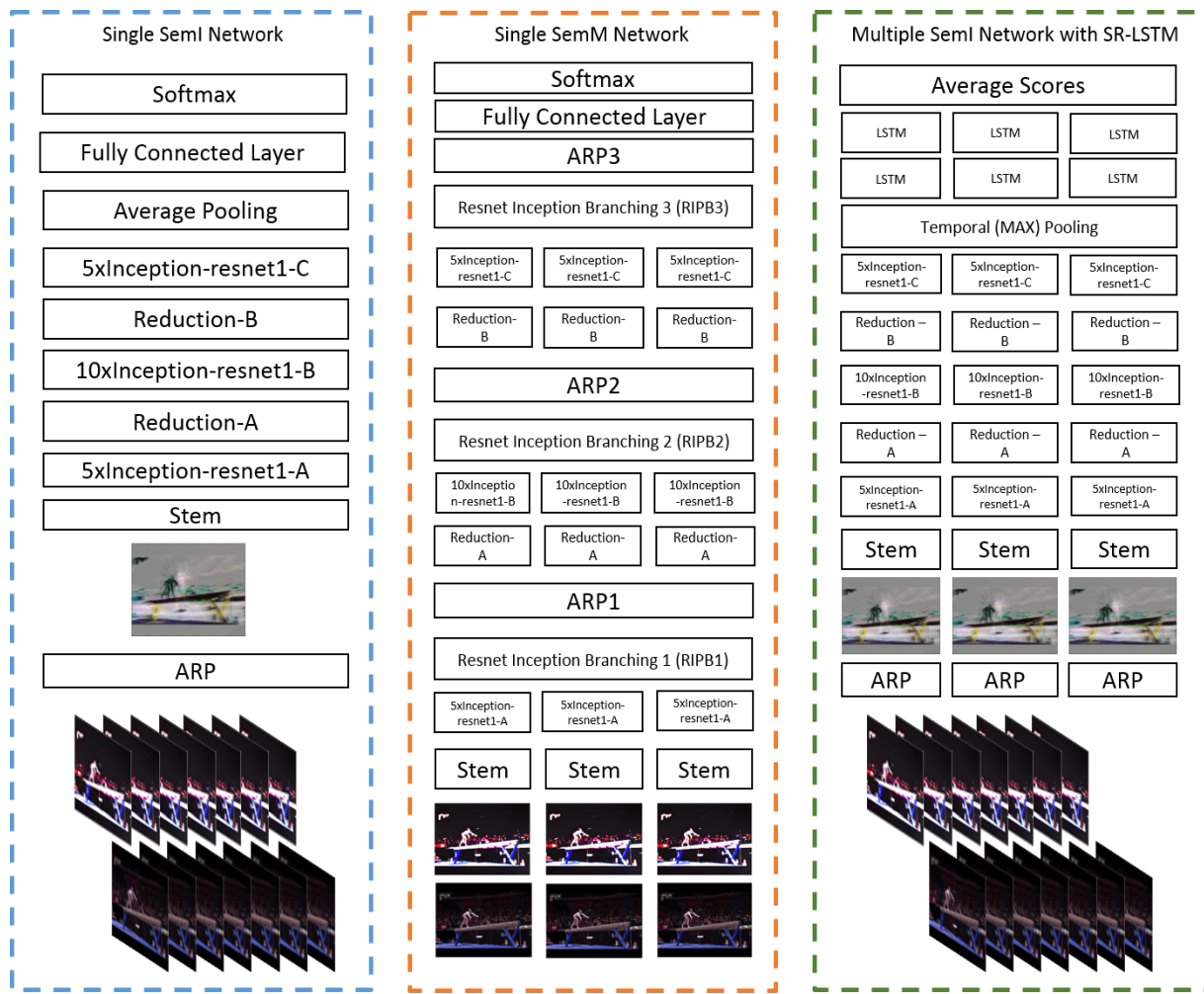

Figure 5 Proposed network architectures for (i) single SemI, (ii) single SemM, and (iii) multiple SemI with SR-LSTM

## 4.2 SR-LSTM

We stack the LSTM layers in sequence with Inception-ResNetv2 to model the temporal dependencies and variances for action recognition. We refer to the stacked network as SR-LSTM. As shown in figure 5, we do not use LSTM stacking for single SemI and single SemM. The reason for not using LSTM is that it cannot leverage the temporal information quite effectively with single SemI and single SemM. Instead, we use SR-LSTM with multiple SemI which proves to be quite effective in terms of modeling the temporal information and variances. The stem of the Inception-ResNetv2 is the same as that of the inceptionv4 network. The detailed composition of stems and inception cells can be found in (Szegedy et al. 2017). The idea for using LSTMs in sequence with CNN architecture was motivated from (Donahue et al. 2017) where they used CNNs for extracting visual features and LSTMs for sequence learning to recognize actions or predict the descriptions for images and videos, accordingly. Similarly, the SR-LSTM can also be used for visual description applications, but in this work, we only focus on action recognition problem. For SemI we segment the RGB frames and apply ARP at the input level that is fed to the Inception-ResNetv2 for training that has been pre-trained on ImageNet ILSVRC 2015. For multiple SemI, we break the video into segments for duration τ and stride $\mathsf{S}$, respectively. The extracted frames from the current segment partially overlap with the previous one which are then fed to SR-LSTM. For each segment of duration τ from which the SemIs are constructed, we apply the temporal i.e., max pooling on the features maps. The features pooled using temporal layer are then fed to the LSTMs which learn the temporally pooled features embedded with ranking scores in a sequential manner. As the frames are scored based on their ranks the temporal variance increase between each frame separated by τ. For instance, the video is divided into multiple clips with certain overlapping for creating multiple SemI, the ranking of a particular frame in one clip may not be the same in the other. This creates temporal variances between the clips which can be efficiently modeled using LSTM layers. The temporal variance concept is one of the keystone for stacking LSTM with Inception-ResNetv2 in a sequential manner.

The SR-LSTM performs the scaling of residual filters before its input to the subsequent connection. The residuals are scaled to stabilize the training process. We observed that as the number of filters increases the residual connections get unstable. The observation is compliant with the studies in (Szegedy et al. 2017) and (He et al. 2016). The original study which proposed residual learning networks (He et al. 2016) suggested using the two-phase training mechanism. This kind of training in its first phase trains the network with very low learning rate whereas the latter phase is trained with high learning rate. However, with SR-LSTM the two-phase learning mechanism was also creating instability as the residual connections were getting unstable, and therefore, the scaling down of residuals was performed before its addition with the activations of a previous layer. It was also observed that the LSTMs also reduces the feature dimensions while learning temporal and spatial features from Inception-ResNetv2 network. The implementation details for SR-LSTM are provided in the experimental and results section.

5. Experiments and Results

In this section, we will present results from our experimental analysis carried out using the proposed video representation and network. This section first reveals the implementation details for all the networks used such as ResNeXt50, ResNeXt101, Inception-ResNetv2, and SR-LSTM. It is well stated that our work is an extension of the representation technique, i.e. dynamic images. In this regard, we first compare the performance of semantic and dynamic images with respect to the activations generated from ResNeXt101's convolutional layers along with the training and validation loss graphs obtained from ResNeXt variants. It will provide an insight as to why the SemI performs better than the dynamic ones. Next, we will provide quantitative analysis on SemI with another competitor in terms of representation, i.e. motion history images. We show that that the SemI performs very well then the motion history images. We then analyze the performance of single SemM by employing approximate rank pooling at different layers and also compare with the performance of single dynamic maps to prove its efficacy. We explain the process of generating multiple SemI to escalate the receptive size of the network by increasing the data magnitude. Similar to the single dynamic maps, we compare the performance of multiple SemI with that of the dynamic ones to show the recognition performance. Optical flows have also been used to represent motion dynamics and have achieved remarkable results as reported in the existing studies. We also generate semantic optical flows (SemOF) and report the classification results on UCF101 and HMDB51 datasets, accordingly. Since we are using different network architectures as compared to the existing studies; it is necessary to compare the performance of all these architectures with each of the given modalities. We compare the performance of SR-LSTM with Inception-ResNetv2 and ResNeXt to show that the proposed architecture can perform well on both the datasets. We present some applications with respect to the proposed representation such as its usage in two-stream, four-stream, and temporal segment networks. Lastly, we will compare our generated results with the state-of-the-art ones.

5.1 Implementation Details

Since we are using various networks for comparison and evaluation; we describe the implementation details of each network one by one followed by the details for RGB and optical flows. All the networks in this study are trained using MATLAB R2018. The training was performed on GPU using Intel Core i7 clocked at 3.4 GHz with 64 GB RAM and NVIDIA GeForce GTX 1070.

ResNeXts: To perform a fair comparison, we use two networks, i.e. ResNeXt50 and ResNeXt101 (Xie et al. 2017) using (32x4d) variant as they have been used for earlier study employing dynamic images. We use the same network settings and protocol for both dynamic and semantic images to perform the comparisons with respect to activations, training loss, and average accuracy.

Inception-ResNetv2 and SR-LSTM: To train Inception-ResNetv2, we first resize our input frames to 299 x 299 which is the input required by the original network. We fine-tuned the network using RMSProp with a decay rate of 0.8 and an epsilon value of 0.9. The learning rate was set to 0.04, decayed after every epoch using an exponential rate of 0.85 and the drop out ratio was set to 0.5. We also used the gradient clipping (Pascanu et al. 2013) to stabilize the training. Since the scaling of residuals is required to overcome the training instability, in this regard, we scale down the residuals with a factor of 0.25 before they are added with the activations of the next layer. We break the Inception-ResNetv2 at $540^{th}$ layer named as 'block17_15' and add a temporal pooling layer to sequentially connect with LSTM as shown in figure 5. We used 2 layers of LSTM with 100 hidden units. We used the ADAM optimizer (Kingma and Ba 2014) with default parameters, i.e. 0.9 and 0.999. The initial learning rate was set to 0.0009 and was decreased by the factor of 0.04 after every epoch.

RGB and Optical flow: The RGB frames are obtained by converting each video to the frame sequences. The warped optical flows using the method (H. Wang and Schmid 2013) were pre-computed, and the flow fields were stored as JPEG images. We rescaled the range of flow values within [0, 255] after clipping with 20 pixels of displacement.

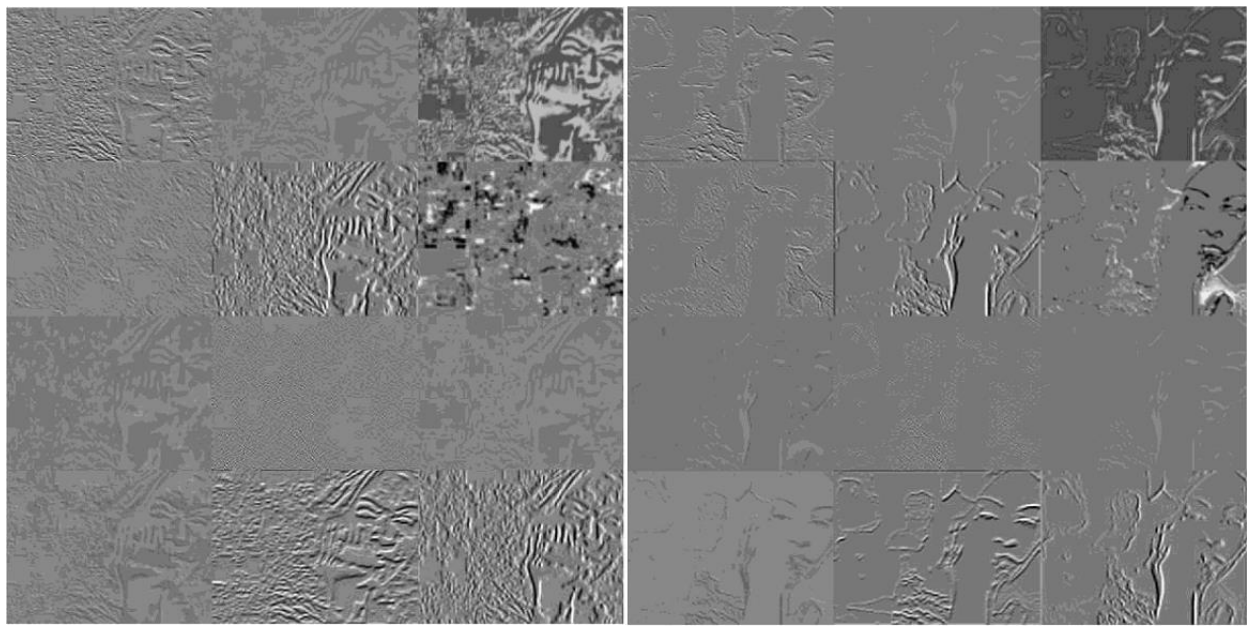

Figure 6 Activations generated for action 'Apply Eye Makeup' from a first convolutional layer of ResNeXt101. Left image generated using a single dynamic image whereas the right one is generated using single SemI.

## 5.2 Single SemI versus Single dynamic image

Our proposed work is an extension to dynamic image construction, so naturally, the question arises why there is a need for SemI? To address this question, we first compare the performance of a single dynamic image with that of single SemI using the activations generated from ResNeXt101's first and third convolutional layer. The activations for the action 'Apply Eye Makeup' are computed from the first convolutional layer and are shown in figure 6. The activations are a good way to qualitatively measure the representations as they exhibit more smooth and less noisy patterns (Fei-Fei et al. 2017; Olah et al. 2017). By visualizing the activations, it is apparent that single SemI generates the better activations as compared to the single dynamic image. The same pattern can also be noticed for the action 'Archery' whose activations are computed from the third convolutional layer as shown in figure 7. After visualizing the difference, we can say that a single SemI generates more smooth and less noisy patterns suggesting that the prior segmentation does improve the action-motion dynamics.

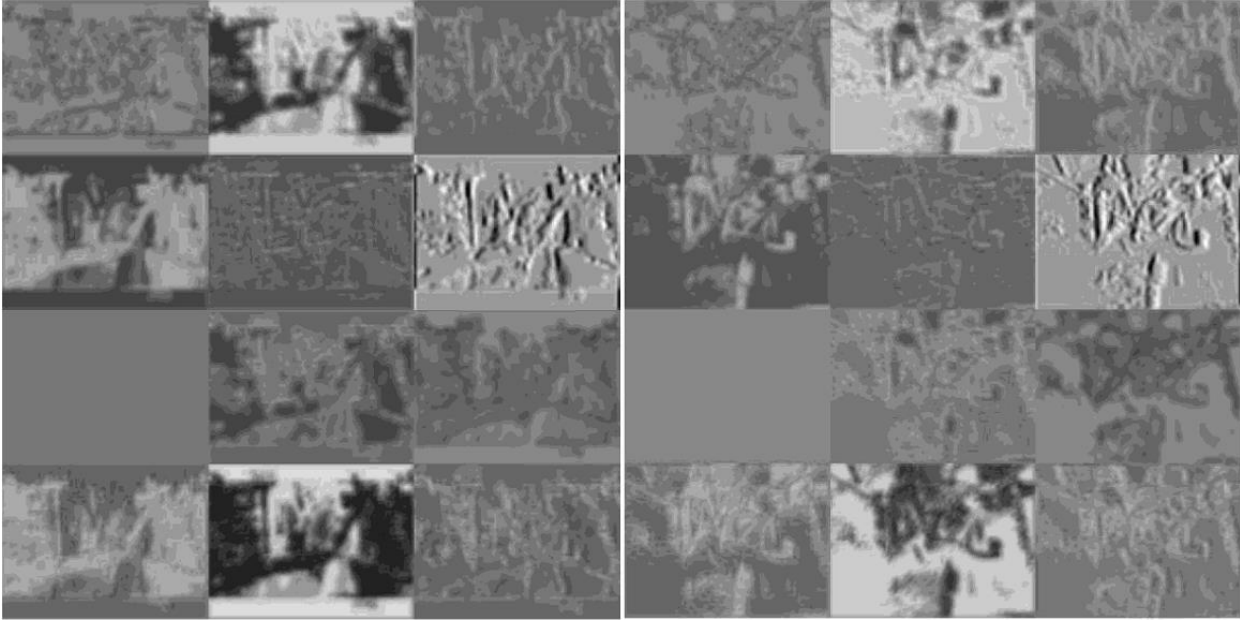

Figure 7 Activations generated for action 'Archery' from the third convolutional layer of ResNeXt101. Left image generated using a single dynamic image whereas the right one is generated using single SemI.

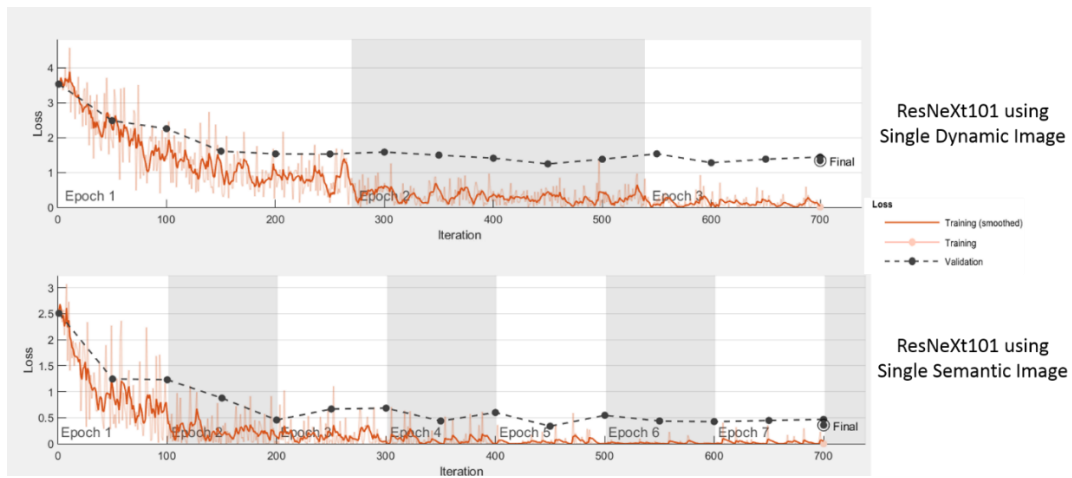

Figure 8 Training and validation loss of single dynamic image and single SemI. From the top (i) Loss from ResNeXt101 using a single dynamic image, and (ii) Loss from ResNeXt101 using single SemI. We can notice a better bias-variance trade-off while using single SemI to that of the single dynamic image.

Next, we compare the single dynamic image and single SemI with the training and validation loss using ResNeXt101 as shown in figure 8. The single SemI shows the trend of faster convergence while yielding low validation error as compared to the single dynamic image. Moreover, the single SemI shows a smaller gap between validation and training loss which is considered to be better bias-variance tradeoff in machine learning theory as compared to its counterpart (Geman et al. 1992; C. Li et al. 2016). We believe that the results prove the effectiveness of a single SemI in comparison to the single dynamic image.

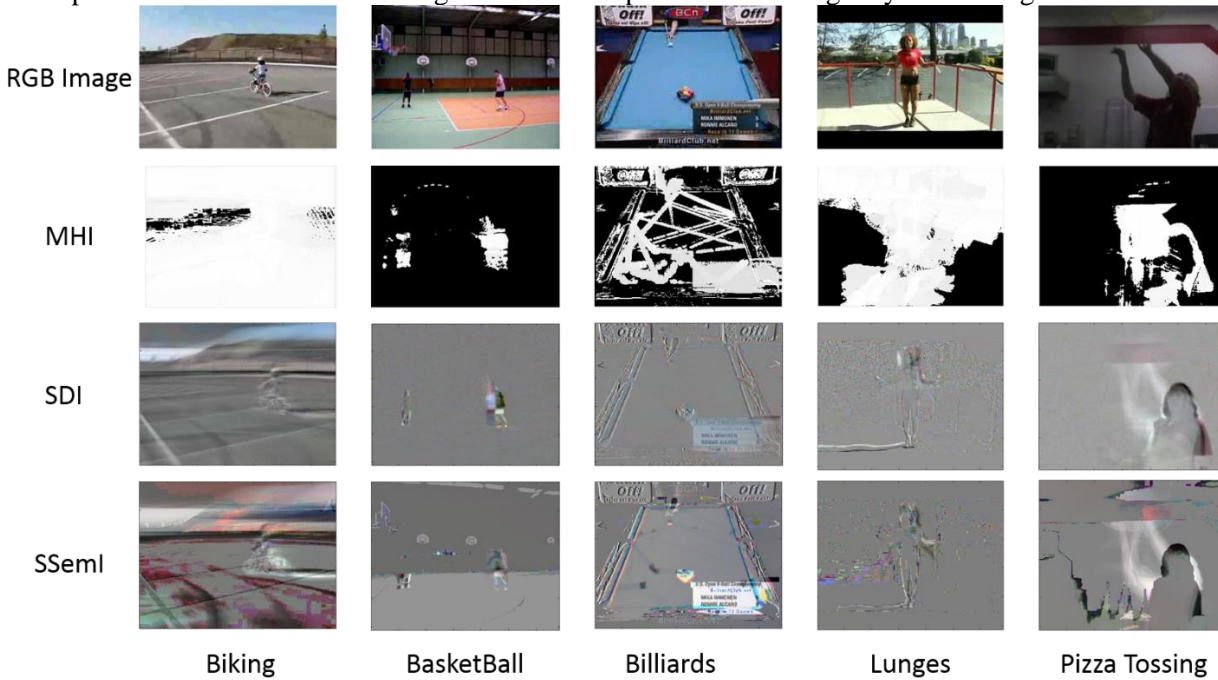

Figure 9 Visualizing different image representations. From top to bottom, RGB image, MHI, single dynamic image, and single SemI. The pixels are mostly bright for MHI images are the motion is performed throughout the frames. The single dynamic image omits out the motion characteristics due to the averaging of pixels which do not change throughout the frames. The single SemI uses prior segmentation which segments different regions as per the motion characteristics to improve the action-motion dynamics. The representations clearly show the difference.

5.3 Single SemI versus existing representations

In this subsection, we provide the qualitative and quantitative analysis for single SemI with the single dynamic image and motion history image. We also compare the results of single SemI with mean and max pooled images which are also considered to be an alternative for generating summarized images (Bilen et al. 2018). The experiments are carried out using ResNeXt50, ResNeXt101, and Inception-ResNetv2 on UCF101 split 1 only. All the representations such as SemI, mean pooled image, max pooled image, motion history image, and single dynamic image are computed offline prior to the training of the networks. We qualitatively compare RGB images, single dynamic image, motion history image, and single SemI in figure 9. The visual difference in the representation can easily be noticed. The motion history image (MHI) fails to represent the motion of actions 'biking' and 'lunges' clearly, as almost all the pixels become white as the motion is present in all regions throughout the frames. MHIs were originally proposed for action recognition, but the pipeline uses moment descriptors which are based on the distance computation for each action category to classify an action. This pipeline worked better for the existing datasets such as KTH and Weizmann, but it is not consistent with the modern datasets. The limitation of the dynamic images can be visualized from 'basketball', 'lunges', and 'pizza tossing' actions as it either fails to represent the motion with respect to the background (see basketball action) or it cannot capture the motion dynamics to its full potential (see lunges and pizza tossing) as compared to the SemI. For instance, in 'lunges' action, the hand movement in single dynamic image is not clear whereas the single SemI maps the motion well enough.

Similarly, for 'pizza tossing' action, the circular motion of hands is quite visible with single SemI as compared to the dynamic image. We present the comparative analysis for MHI, mean pooled, max pooled, single dynamic image, and single SemI, in table 1, respectively.

Table 1 Comparison of MHI, mean pooled image, max pooled image, single dynamic image, and single SemI with respect to the mean class accuracy on UCF101 split 1.

| Image Representation | Architecture | Accuracy |
|---|---|---|
| Single MHI | ResNeXt50 | 58.8% |
| Single MHI | ResNeXt101 | 60.1% |
| Mean Image | ResNeXt50 | 64.8% |
| Mean Image | ResNeXt101 | 65.6% |
| Max Image | ResNeXt50 | 59.9% |
| Max Image | ResNeXt101 | 61.3% |
| Single dynamic image | ResNeXt50 | 70.9 % |
| Single dynamic image | ResNeXt101 | 71.6 % |
| Single SemI | ResNeXt50 | 72.2% |
| Single SemI | ResNeXt101 | 73.1% |
| **Single SemI** | **Inception-ResNetv2** | **74.7%** |

The quantitative results show that the single SemI achieved 13.0%, 7.5%, 11.8%, and 1.5% gain inaccuracies over single MHI, mean, max, and single dynamic image using ResNeXt101, respectively. The result proves our assumption that the use of prior segmentation and addition of semantic information (background) can help in modeling better action-motion dynamics as compared to the existing representations. It is further noticed that when applying Inception-ResNetv2, we can boost the accuracy further by 2.5% and 1.6% in comparison to the ResNeXt architectures.

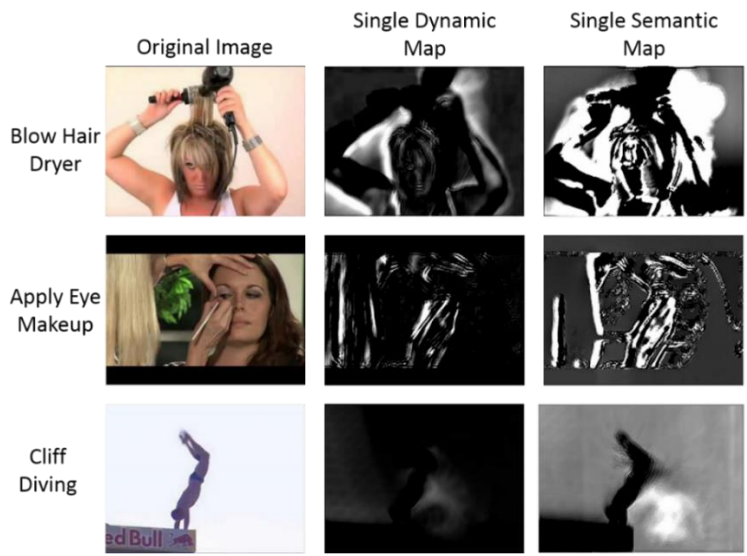

Figure 10 Visualization of RGB and segmented images when applied ARP after the first convolution layer of ResNeXt101

5.4 Single Semantic Map

In the previous subsection, we used single SemI which was generated using approximate rank pooling before feeding them to the deep learning architecture. To generate the single SemM, we move the approximate rank pooling layer deeper in the architecture which also proves the flexibility of the representation method. We first compare the single dynamic map (Bilen et al. 2016) using ResNeXt101 and compare it with the single SemM to perform a fair comparison. We also compare the performance of

ARP at different branching factors of Inception-ResNetv2 architecture (see figure 5). The visualization of the single dynamic map and single SemM is shown in figure 10 which are acquired using ResNeXt101. The strength of SemM is clearly advocated by visualizing the maps of 'cliff diving' action where the single dynamic map fails to capture any motion characteristic.

Table 2 Mean class accuracy for single dynamic maps and single SemM for UCF101 and HMDB51 datasets

| Method | Architecture | UCF101 | HMDB51 |
| --- | --- | --- | --- |
| Single-dynamic map | ResNeXt50 | 68.9% | 39.7% |
| Single-dynamic map | ResNeXt101 | 69.2% | 39.8% |
| Single SemM | ResNeXt50 | 71.3% | 40.6% |
| Single SemM | ResNeXt101 | 72.5% | 42.2% |
| Single SemM (RIPB1) | Inception-ResNetv2 | 73.8% | 44.8% |
| Single SemM (RIPB2) | Inception-ResNetv2 | **75.9%** | 46.5% |
| Single SemM (RIPB3) | Inception-ResNetv2 | 75.6% | **47.1%** |

The comparison of mean class accuracy on UCF101 and HMDB51 datasets using a single dynamic map and single SemM is reported in table 2. We show that by using single SemM, we can achieve better recognition results than that of the single dynamic map. The single SemM can achieve better recognition results than that of the single dynamic map when using either of ResNeXt architecture. For quantitative results using Inception-ResNetv2, we only use the single SemM. We achieve the best accuracy of 75.9% while breaking the network at RIPB 2 and 47.1% on HMDB51 by breaking the network at RIPB 3. Considering the margin of gain we get, i.e. 6.7% and 7.3% on UCF101 and HMDB51; we can say that the single SemM improves the recognition results as well as benefits the boost in accuracy with RIPB using Inception-ResNetv2.

5.5 <u>Multiple Semantic Images</u>

Although it is presented in quantitative results that single SemI performs better than the existing video representations which summarize the motion in single images the accuracy is still not enough to compete with state-of-the-art methods. One reason for not attaining higher accuracy is the lack of annotated data needed to fine tune the network. Unfortunately, UCF101 and HMDB51 datasets have few videos for each category, therefore extracting multiple SemI could help in overcoming the said problem. The extraction of multiple SemI is accomplished by breaking each video in clips, i.e. duration of frames. In this regard, each video is divided into multiple clips with partially overlapping frames. It can be also be considered as a data augmentation step where we increase the volume of data by dividing the original video to a number of clips with duration $\tau$ and stride $\mathfrak{S}$. For network architecture such as ResNext, and Inception-ResNetv2 we add the temporal pooling layer to merge all the subsequences into one for further classification. We use max pooling as our temporal pooling layer for both the architectures due to the best reported accuracy in (Bilen et al. 2018). In order to show that the use of multiple SemIs can model better action-motion dynamics for complex actions and many temporal changes, we visualize the SemI with different window sizes in figure 11. It can be visually noticed that for the periodic and longer motions such as 'biking' action, the multiple SemI stretches the same motion dynamics onto the subsequent frames which introduces the motion blur kind of effect. The actions such as 'hula hoop' benefits from the long term motion dynamics yet the motion is quite well preserved even with 10 frames. We can also notice a disadvantage of using the whole video length for summarization as depicted in 'pommel horse' action. The single SemI does not capture the complex motion pattern for this action. Due to the repetition of same action over and over, the motion is averaged out. However, using fewer frames, i.e., 50 or 10, can reduce the said limitation.

In this regard, we will first perform analysis using different window sizes and strides as shown in table 3. We computed the mean class accuracy for UCF101's first split and noticed that the best accuracy is achieved by using the duration of 15 with 40% overlap. It is also interesting to see that while increasing the window size the accuracy starts decreasing which proves our assumption for 'pommel horse' action illustrated in figure 11. For our further experiments, we will use $\tau = 15$ and $\mathfrak{S} = 9$, respectively.

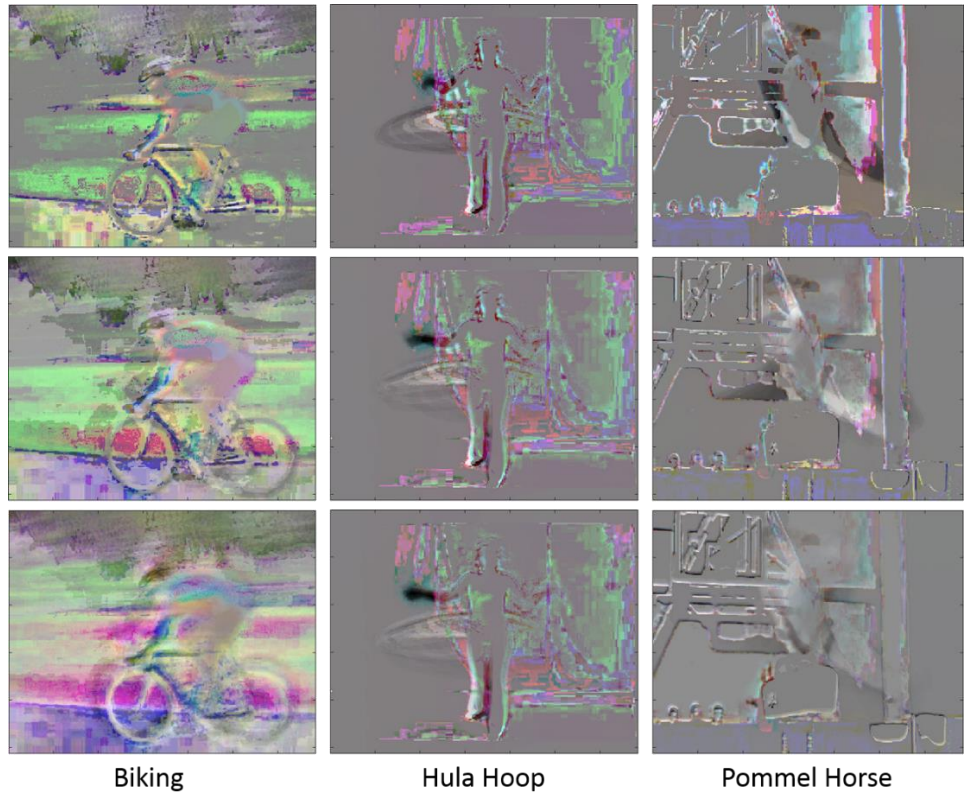

Figure 11 Visual Analysis of multiple SemIs with respect to the varying window sizes. The top row shows the SemI for window size 10, the middle row comprises of images with window size 50, and the bottom row depicts the single SemI.

Table 3 Effect of the window size and overlapping on mean class accuracy for UCF101.

| Network Architecture | Window τ | Stride $\varsigma$ | Accuracy |
|---|---|---|---|
| ResNeXt101 | 5 | 3 | 83.5% |
| Inception-ResNetv2 | 5 | 3 | 83.9% |
| ResNeXt101 | 10 | 2 | 86.4% |
| Inception-ResNetv2 | 10 | 2 | 86.7% |
| ResNeXt101 | 10 | 4 | 87.2% |
| Inception-ResNetv2 | 10 | 4 | 87.6% |
| ResNeXt101 | 10 | 6 | 87.8% |
| Inception-ResNetv2 | 10 | 6 | 88.0% |
| ResNeXt101 | 15 | 6 | 88.2% |
| Inception-ResNetv2 | 15 | 6 | 88.7% |
| ResNeXt101 | 15 | 9 | 88.0% |
| **Inception-ResNetv2** | **15** | **9** | **88.9%** |
| ResNeXt101 | 20 | 6 | 80.1% |
| Inception-ResNetv2 | 20 | 6 | 80.6% |

We now compare the recognition accuracy of multiple SemI with multiple dynamic images and RGB images in table 4. It is observed that the Inception-ResNetv2 performs better than the ResNeXt101 which is compliant to the ILSVRC 2017 imageNet results where Inception-ResNetv2 also performs better i.e., Top 1 error of 19.9, as compared to the ResNeXt101 yielding an error of 21.2, respectively (Hu et al. 2017). We observed that the multiple SemI attain the performance at par with RGB static images for UCF101 but achieved recognition accuracy of 59.6% on HMDB51which is 5.0% better than that of static RGB images.

We believe that the reason for the better gain is the ability of multiple SemIs to represent more intricate action-motion dynamics and to leverage the semantic information better due to the complexity of backgrounds in HMDB51 dataset.

Table 4 Mean class classification accuracy for UCF101 and HMDB51 datasets. Table also shows comparative analysis for RGB static images, multiple dynamic images, and multiple SemIs using ResNeXt and Inception-ResNetv2 architectures.

| Method | Architecture | UCF101 | HMDB51 |
| --- | --- | --- | --- |
| RGB Static Images | ResNeXt101 | 87.9% | 53.8% |
| RGB Static Images | Inception-ResNetv2 | 88.6% | 54.6% |
| Multiple Dynamic Images | ResNeXt101 | 87.1% | 57.7% |
| Multiple Dynamic Images | Inception-ResNetv2 | 87.6% | 58.2% |
| Multiple SemI | ResNeXt101 | 87.8% | 59.4% |
| **Multiple SemI** | **Inception-ResNetv2** | **88.9%** | **59.9%** |

### 5.6 Warped and Semantic Optical Flows

As illustrated in previous subsections, the ARP transforms the RGB images to semantic images, i.e. low-level to mid-level motion representation. The optical flows are another example of mid-level motion representations which have been used extensively in terms of two-stream networks. It was proved in (Bilen et al. 2018) that applying the ARP to optical flows transforms the mid-level representation to high-level which can improve the recognition results. To this extent, we apply the ARP to the warped optical flows from the segmented images overlaid with background information to generated semantic optical flows (SemOF). Similar to the previous analysis, we apply ARP to 15 optical flow frames for generating SemOF. We first visualize the warped optical flows and semantic optical flows in figure 12 to analyze the difference qualitatively. It can be visualized that the SemOF capture long-term motion dynamics as compared to the warped optical flows. This is because the warped optical flows can only calculate the motion patterns between subsequent frames — for instance, the SemOF for action 'boxing punching bag' and 'billiards' exhibit history of temporal actions for a longer span as compared to the warped optical flows. Furthermore, the SemOF can represent the direction of the velocity for the optical flows as shown for 'boxing punching bad' and 'wall pushups,' the direction is horizontal whereas the 'brushing teeth' and 'clean and jerk' the direction is vertical.

We also compare the performance of warped and dynamical optical flows with the SemOF quantitatively to prove its efficacy. The SemOF achieved 2.0%, 0.2%, and 3.3%, 0.2%, better accuracy on UCF101 and HMDB51 than warped and dynamic optical flows, respectively. These results were obtained using ResNeXt101; similar trend is observed when using Inception-ResNetv2 where the SemOF improved the accuracy up to 2.3% and 0.4% on UCF101 for warped and dynamic optical flows. The gain in accuracy for HMDB51 was noted to be 3.6% and 0.3% from warped and dynamic optical flows using Inception-ResNetv2. The results prove that the transition from mid-level to high-level motion representation improves the recognition performance.

Table 5 Mean class accuracy for UCF101 and HMDB51 datasets using warped, dynamic, and semantic optical flows.

| Method | Architecture | UCF101 | HMDB51 |
| --- | --- | --- | --- |
| Warped Optical Flows | ResNeXt101 | 85.1% | 56.0% |
| Warped Optical Flows | Inception-ResNetv2 | 85.3% | 56.3% |
| Dynamic Optical Flows | ResNeXt101 | 86.9% | 59.1% |
| Dynamic Optical Flows | Inception-ResNetv2 | 87.2% | 59.6% |
| SemOF | ResNeXt101 | 87.1% | 59.3% |
| **SemOF** | **Inception-ResNetv2** | **87.6%** | **59.9%** |

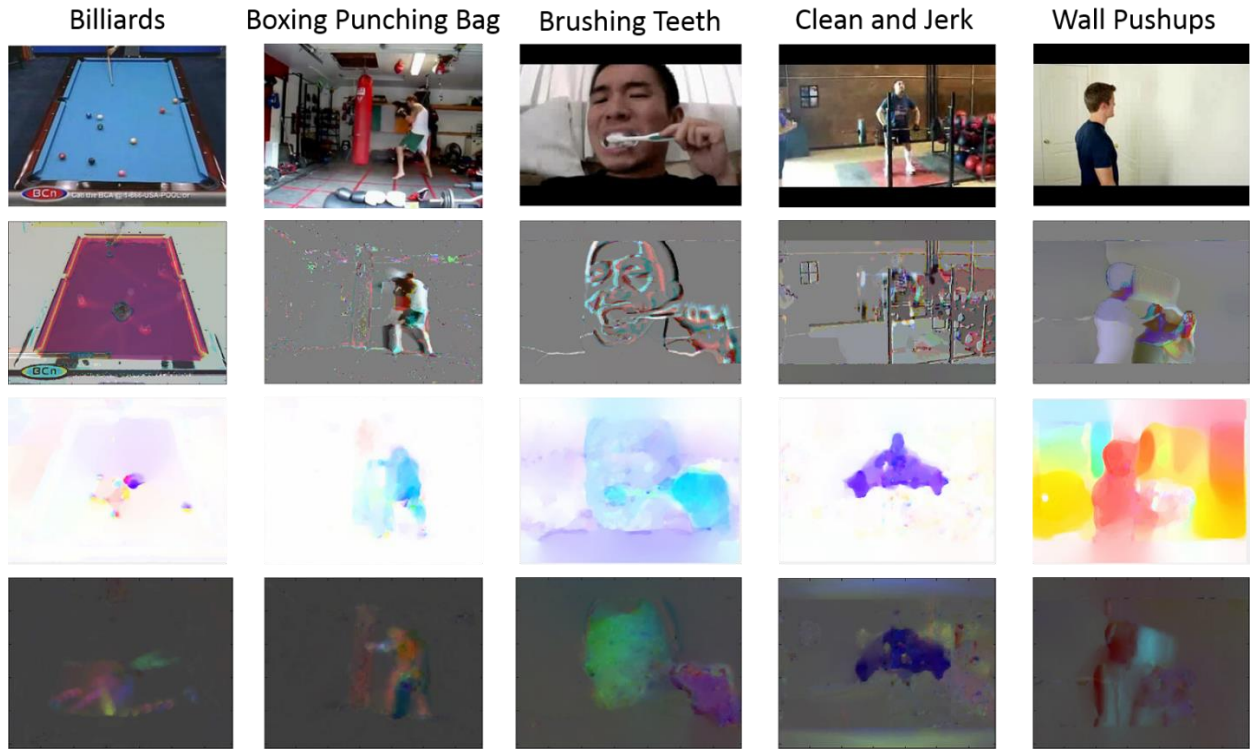

Figure 12 Visualizing the static RGB images in the first row, SemI in the second row, warped optical flows in the third row, and SemOF in the fourth row.

5.7 SR-LSTM versus Inception-ResNetv2 and ResNeXts

So far, we have used two deep networks, i.e. Inception-ResNetv2 and ResNeXts. We showed that the Inception-ResNetv2 could achieve better results than the ResNeXts. Considering that both the networks extract different features which are illustrated by the frequency responses obtained using the first convolutional layer as shown in figure 13, we assume that the features extracted Inception-ResNetv2 along with its deeper architecture. The frequency responses were obtained by passing the same RGB image and applying the max pooling to select the filter activation with maximum values. We then plot the frequency responses of the green channel of the selected activation The ResNeXt101 weight show the bandpass filter in the centers for a particular direction surrounded by the low pass filters in other direction. With respect to the signal processing analogy, we realize that the filter's cut-off the upper half of the frequency range for most of the signal which is similar to the antialiasing filter. On the other hand, the frequency response of Inception-ResNetv2's max pooled image shows two high pass filters appearing on the horizontal edges while small magnitude low pass filters are present on the extreme ends. We assume that the low pass filters smooth the edges while the intensities in a particular range of two scales are computed for feature maps.

We know the SemI use ARP which retains the temporal information from the video sequences by ranking them. In this regard, we assume that by stacking LSTM network in sequence with Inception-ResNetv2 the recognition results can be improved due to the learning of temporal variances introduced when dividing the video into different windows. We refer to this network as SR-LSTM. To prove the validity of our assumption, we carried out experiments on multiple dynamic images, multiple SemI, warped optical flows, dynamic optical flows, and SemOF using SR-LSTM and computed the mean class accuracy as shown in table 6. The trend shows that the SR-LSTM can improve the recognition performance for all the modalities used for action recognition. It is to be noted that all of these modalities somehow incorporate temporal information, let it be ARP or motion information from warped optical flows.

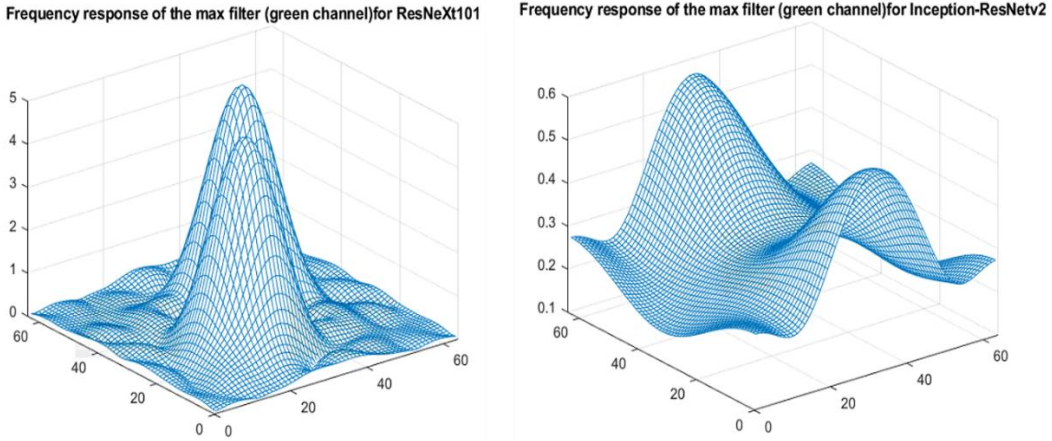

Figure 13 Visualizing the frequency response of the maximum filters using green channels for ResNeXt101 and Inception-ResNetv2 networks.

Table 6 Comparative analysis for UCF101 and HMD51 using ResNeXt101, Inception-ResNetv2, and SR-LSTM networks.

| Method | Architecture | UCF101 | HMDB51 |
|---|---|---|---|
| Multiple Dynamic Images | ResNeXt101 | 87.1% | 57.7% |
| Multiple Dynamic Images | Inception-ResNetv2 | 87.6% | 58.2% |
| Multiple Dynamic Images | SR-LSTM | 88.9% | 60.1% |
| Multiple Semantic Images | ResNeXt101 | 87.8% | 59.4% |
| Multiple Semantic Images | Inception-ResNetv2 | 88.2% | 59.9% |
| Multiple Semantic Images | SR-LSTM | 90.3% | 62.9% |
| Warped Optical Flows | ResNeXt101 | 85.1% | 56.0% |
| Warped Optical Flows | Inception-ResNetv2 | 85.3% | 56.3% |
| Warped Optical Flows | SR-LSTM | 86.3% | 56.8% |
| Dynamic Optical Flows | ResNeXt101 | 86.9% | 59.1% |
| Dynamic Optical Flows | Inception-ResNetv2 | 87.2% | 59.6% |
| Dynamic Optical Flows | SR-LSTM | 88.5% | 61.3% |
| Semantic Optical Flows | ResNeXt101 | 87.2% | 59.3% |
| Semantic Optical Flows | Inception-ResNetv2 | 87.8% | 60.0% |
| Semantic Optical Flows | SR-LSTM | 89.8% | 62.4% |

We assume that SR-LSTM can leverage the temporal information for improving the recognition holds since the recognition results are boosted for all the modalities. The multiple dynamic images see 1.8% and 1.3% growth from ResNeXt101 and Inception-ResNetv2 networks. The gain for multiple SemI is noted to be 2.5% and 2.1% compared to the other architectures. Warped optical flows record 1.2 and 1.0% gain whereas the dynamic optical flows show 1.6% and 1.3% gain in accuracy. All these gains are reported from UCF101 dataset. The similar trends are observed for HMDB51 as well. The highest gain for HMDB51 was observed for multiple SemI gaining 3.5% and 3.2% accuracy over ResNeXt101 and Inception-ResNetv2. The results prove the effectiveness of adding LSTM in sequence with Inception-ResNetv2 to leverage the temporal information and temporal variance for better recognition performance.

5.8 Two- and Four- Stream Networks

It is well established throughout section 2 that the researchers use multi-stream networks for boosting the recognition performance by combining the results of networks trained using different modalities. We performed the analysis for multi-stream networks using the combination of modalities such as RGB, multiple SemI, warped optical flows and SemOF. Each of the streams in multi-stream networks is trained separately for different modalities, but their results are combined using average scores for each class. The final classification label will be drawn by selecting the class having a maximum average score. Our proposed multi-stream networks, i.e. two- and four-stream network, are depicted in figure 14. The analysis for late fusion is performed using SR-LSTM network as it achieved the best results for recognition on both datasets. The mean class accuracies for UCF101 and HMDB51 datasets using multi-stream networks are presented in table 7. The best results are achieved using multiple SemI and SemOF using a two-stream variant which gives a boost of 6.1%, 4.4%, 8.4%, and 4.9% accuracy over individual modalities of static RGB, multiple SemI, warped optical flows, and SemOF on UCF101. Similarly, the gain is noted to be 10.2%, 7.2%, 13.3%, and 7.7% better than the individual modalities of static RGB images, multiple SemI, warped optical flows, and SemOF on HMDB51. Finally, the maximum accuracy was achieved using a four-stream variant which gives a boost of 1.2% and 3.4% over two-stream variant for UCF101 and HMDB51 datasets, respectively.

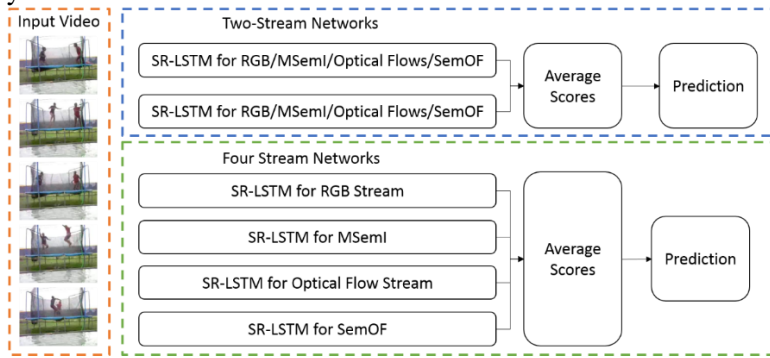

Figure 14 Illustration of multi-stream network architecture using our proposed video representations

Table 7 Comparative analysis for a combination of various modalities with respect to the mean class accuracy on UCF101 and HMDB51.

| Combination | UCF101 | HMDB51 |
|---|---|---|
| Static RGB + Multiple SemI | 92.8% | 64.6% |
| Static RGB + Warped Optical Flows | 94.2% | 68.1% |
| Multiple SemI + Warped Optical Flows | 94.4% | 67.9% |
| Static RGB + SemOF | 93.9% | 67.8% |
| Warped Optical Flows + SemOF | 92.1% | 65.8% |
| **Multiple SemI + SemOF** | **94.7%** | **70.1%** |
| **Static RGB + Multiple SemI + Warped Optical Flows + SemOF** | **95.9%** | **73.5%** |

5.9 Temporal Segment Networks

The temporal segment network has proved that using good practices for making deep architectures learn can improve the recognition results for applications such as action recognition. The main characteristics of temporal segment networks include a sparse sampling of short snippets from the video and the consensus among the snippets for drawing out the predictions. The temporal segment networks have been used for two and three modalities in their original work (L. Wang, Xiong, et al. 2016). In this paper, we use the same architecture and method to train the network, but instead of using RGB and optical flows we use multiple SemI and SemOF, respectively. To match the number of samples we only consider the last two frames in a window to generate the warped optical flows. The comparative analysis for different modalities using temporal segment networks is shown in table 8. The results show that the three modalities with temporal segment networks achieve better results than our two-stream networks on both the datasets.

Table 8 Comparative analysis for two and three input modalities using temporal segment networks. The () shows the credits for different CNN to determine the consensus from a specific modality.

| Combination (Ratio) | UCF101 | HMDB51 |
|---|---|---|
| Static RGB + Warped Optical Flow (1:1.5) | 93.1% | 67.6% |
| Multiple SemI + Warped Optical Flow (1.5:1) | 94.2% | 68.7% |
| Multiple SemI + Semantic Optical Flow (1:1.5) | 94.7% | 69.5% |
| Multiple SemI + Semantic Optical Flow + Warped Optical Flow (1:1:0.5) | **95.2%** | **72.1%** |

Table 9 Comparison of mean class accuracy with state-of-the-art methods without the fusion of IDTs

| Methods | HMDB51 | UCF101 |
|---|---|---|
| C3D+SVM (Tran et al. 2015) | - | 85.2% |
| 2S-CNN (Joe Yue-Hei Ng et al. 2015) | 59.4% | 88.0% |
| 2S-CNN+Pool (Joe Yue-Hei Ng et al. 2015) | - | 88.2% |
| 2S-CNN+LSTM (Joe Yue-Hei Ng et al. 2015) | - | 88.6% |
| TDD (L. Wang et al. 2015) | 60.2% | 90.3% |
| Objects + Motion(R*) (Jain et al. 2015) | 61.4% | 88.5% |
| FSTCN (Sun et al. 2015) | 59.1% | 88.1% |
| Comp-LSTM (Srivastava et al. 2015) | 44.0% | 84.3% |
| CNN-hid6 (Zha et al. 2015) | - | 79.3% |
| MoFAP (L. Wang, Qiao, et al. 2016) | 61.7% | 88.3% |
| KVMF (Zhu et al. 2016) | 63.3% | 93.1% |
| Tube ConvNet (Z. Li et al. 2016) | - | 92.3% |
| Sialency Context (Chen et al. 2016) | - | 92.4% |
| SR-CNN (Yifan Wang et al. 2016) | - | 92.6% |
| Temporal Segment Networks (L. Wang, Xiong, et al. 2016) | 69.4% | 94.2% |
| Two Stream DIN (CaffeNet) (Bilen et al. 2016) | 42.8% | 76.9% |
| Two Stream I3D (Carreira and Zisserman 2017) | 66.4% | 93.4% |
| ST-Pyramid Networks (Yunbo Wang et al. 2017) | 68.9% | 94.6% |
| LTC (Varol et al. 2018) | 64.8% | 91.7% |
| Two Stream DIN (ResNeXt50) (Bilen et al. 2018) | 67.5% | 93.9% |
| TLE (Diba et al. 2017) | 71.1% | 95.6% |
| TS-LSTM (Ma et al. 2018) | 69.0% | 94.1% |
| Two-Stream I3D (Carreira and Zisserman 2017) + (Kinetics 300k) | 80.7% | 98.0% |
| Four Stream DIN (ResNeXt50) (Bilen et al. 2018) | 71.5% | 95.0% |
| Four Stream DIN (ResNeXt101) (Bilen et al. 2018) | 72.5% | 95.5% |
| ***Two-Stream Networks (SemIN)*** | ***70.1%*** | ***94.7%*** |
| ***Temporal Segment Networks (multiple SemI + semantic optical flows + warped optical flows) 1:1:0.5*** | ***72.1%*** | ***95.2%*** |
| ***Four Stream Networks (SemIN)*** | ***73.5%*** | ***95.9%*** |

5.10 State-of-the-art methods

In this section, we compare our experimental results with the state-of-the-art methods for action recognition using videos. We mentioned in section 2 that most of the existing studies perform late fusion with improved dense trajectories (IDTs) to boost their recognition accuracy. In this regard, we divide the state-of-the-art results into two categories. The first includes the results without the use of IDTs so that the method could be judged on its intrinsic characteristics. The second, presents the recognition accuracies from the methods when combined with IDTs, accordingly. We assume that such categorization of results for comparison would be fair enough. We present the results without combining IDTs in table 9. It can be noticed that our two-stream networks outperform many state-of-the-art methods. The temporal segment networks with proposed representations not only improves the accuracy from the original study but also achieves better results than many existing works. Temporal linear encoding method (Diba et al. 2017) is the only method which performs at par with the improved temporal segment networks on UCF101 but could not generalize the same performance on HMDB51 dataset. We achieved the best results using four-stream networks by attaining 95.9% and 73.5% on UCF101 and HMDB51 dataset. The proposed four-stream network improves the recognition accuracy by 0.4% and 1.0% for UCF101 and HMDB51 from the one proposed in (Bilen et al. 2018). We outperform all the state-of-the-art methods except I3D (Carreira and Zisserman 2017) + kinetics 300k. The reason for the high accuracy of this method is that it was pre-trained on additional 300k kinetic sequences while relaying on two-stream networks. It should also be noticed that when the I3D method was only trained on HMDB51 and UCF101, even our two-stream networks outperformed their method. It implies that we can also use I3D + kinetics 300k pre-trained network with multiple SemI and SemOF to boost the accuracy as we did with temporal segment networks.

We also perform the fusion of two-stream and four-stream networks with IDTs for comparison with state-of-the-art works as reported in table 10. It is surprising that the results from our four-stream networks outperform existing state-of-the-art methods combined with IDTs. The result implies that the improvement in accuracy is mostly due to the intrinsic characteristics of our video representation and network architecture. The combination of our four-stream networks with IDTs provides 1.5% and 3.3% boost of accuracy on UCF101 and HMDB51, respectively.

Table 10 Comparison of mean class accuracy with state-of-the-art methods when combined with IDTs

| Methods | HMDB51 | UCF101 |
|---|---|---|
| FV + IDT (Perronnin et al. 2010) | 57.2% | 84.8% |
| SFV + STP + IDT (Perronnin et al. 2010) | 60.1% | 86.0% |
| FM + IDT (Peng et al. 2014) | 61.1% | 87.9% |
| MIFS + IDT (Laptev 2005) | 65.1% | 89.1% |
| CNN + hid6 + IDT (Zhao and Pietikainen 2007) | - | 89.6% |
| C3D Ensemble + IDT (Tran et al. 2015) | - | 90.1% |
| C3D + SVM + IDT (Tran et al. 2015) | - | 90.4% |
| TDD + IDT (L. Wang et al. 2015) | 65.9% | 91.5% |
| Sympathy (de Souza et al. 2016) | 70.4% | 92.5% |
| Two-stream Fusion + IDT (Feichtenhofer, Pinz, and Zisserman 2016) | 69.2% | 93.5% |
| ST-ResNet + IDT (Feichtenhofer, Pinz, and Wildes 2016) | 70.3% | 94.6% |
| LTC + IDT (Varol et al. 2018) | 67.2% | 92.7% |
| Four Stream + IDT with ResNeXt50 (Bilen et al. 2018) | 74.2% | 95.4% |
| Four Stream + IDT with ResNeXt101 (Bilen et al. 2018) | 74.9% | 96.0% |
| ***Two-Stream Networks (SemIN) + IDT*** | ***72.3%*** | ***95.7%*** |
| ***Four Stream Networks (SemIN) + IDT*** | ***76.8%*** | ***97.4%*** |

5.11    Discussion

In this section, we provide some insights with respect to the experimental analysis. It was found that the best relative performance when using multiple SemI was obtained for 'Nun Chucks' and 'Pull Up' action on UCF101 and HMDB51, respectively. However, when compared with static RGB images, the best accuracy improvement was obtained for 'High Jump' action on UCF101 while on the HMDB51 the best improvement was recorded for 'Sommer Sault' and 'throw' action, respectively. We also found that some 'Pizza Tossing' and 'Salsa Spin' actions were confused with 'Nun Chucks' due to the similar circular motions, which are interestingly very different actions as per their characteristics. Furthermore, as pointed out in (Bilen et al. 2018), the 'Breast Stroke' was confused with 'Rowing' and 'Front Crawling' actions. For the two-stream networks the actions which achieved the best relative performance on HMDB51 were

'Pull Up,' 'Ride Bike,' and 'Golf' whereas for the UCF101 the actions were 'Nun Chucks' and 'Jumping Jack.' The four-stream network proposed in (Bilen et al. 2018) suggested the most challenging actions be 'Pizza Tossing,' 'Lunges,' 'Hammer Throw,' 'Shaving Beard', and 'Brushing Teeth' with their respective accuracies of 74, 74.6, 77.2, 78.7 and 80.2 %. Our proposed four-stream network improved the accuracy for each of these actions by 2.6%, 6.4%, 15.6%, 7.8%, and 0.3%, respectively. The most challenging actions on HMDB51 were found to be 'Sword' and 'Wave.' We assume that the similar motion characteristics may be the reason due to which certain actions are confused with one another. However, if combined with the pose analysis or facial landmarks, the SemI may overcome the shortcomings mentioned above.

6. Conclusion and Future Work

In this paper, we propose an improvement to work presented in (Bilen et al. 2018) by segmenting the image and overlaying it with a static background. Such pre-processing method for adding semantics not only improves the action-motion dynamics but also is helpful in mapping semantic information for many actions which are performed at a particular location such as 'Basketball,' 'Breast Stroke,' and more. We also exploit the design choices for computing semantic maps across intermediate layers of Inception-ResNetv2 to prove the flexibility of the representation method. We present the SR-LSTM network which is the result of sequentially combining the base network, i.e. Inception-ResNetv2 with LSTMs. The experimental results show that LSTMs help to model the temporal dynamics from the extracted features by the network across time. The use of temporal information encoded by approximate rank pooling helps LSTMs to learn temporal dependencies and temporal variances to improve the recognition performance. It is reflected by the mean class accuracy on both UCF101 and HMDB51 datasets. Our two- and four-stream variants of SR-LSTM networks show promising results and achieve state-of-the-art performance. The maximum accuracy we achieved so far is by combining our four-stream networks with IDTs, i.e. 97.4% and 76.8% on UCF101 and HMDB51. The modalities were also trained using temporal segment networks which elicited a positive trend towards the improvement in recognition performance. It can simply be assumed that using the SemI and SemOF in the well-known network pipelines such as Two-stream I3D combined with kinetics 300k data, can generate better results than the original variant.

We already discussed some the limitations with respect to the confusing classes in the previous section. Moreover, the SemI and SemOF do not map the motion dynamics well for many complex actions such as 'Pommel Horse' or the actions with abrupt motions such as 'Cricket Shot' and 'Parallel Bars.' Furthermore, the SemOF reveal similar motion characteristics for 'Shaving Beard,' 'Brushing Teeth,' and 'Applying Lipstick' which does not contribute to improving the recognition performance. We believe that by adding another semantic information such as objects, can improve the recognition performance where the objects are distinguishable for each of the above-mentioned actions. In future, we want to explore the use of SemI for cross domain and transfer learning approaches by considering the mutual actions in both datasets such as 'Golf,' 'Dive,' 'Fencing,' 'Jump,' 'Pull Up,' and more.

Acknowledgement

This research was supported by Basic Science Research Program through the National Research Foundation of Korea (NRF) funded by the Ministry of Education (2018R1D1A1B07049113).

Conflict of Interest

The authors declare that they have no conflict of interest